# Tensor-based process control and monitoring for semiconductor manufacturing with unstable disturbances

Yanrong Li[a], Juan Du[b,c*], Fugee Tsung[d,e], Wei Jiang[a]

[a]*Antai College of Economics and Management, Shanghai Jiao Tong University, Shanghai, China;* [b]*Smart Manufacturing Thrust, Systems Hub, The Hong Kong University of Science and Technology (Guangzhou), Guangzhou, China;* [c]*Department of Mechanical and Aerospace Engineering, The Hong Kong University of Science and Technology, Hong Kong SAR, China;* [d]*Department of Industrial Engineering and Decision Analytics, The Hong Kong University of Science and Technology, Hong Kong SAR, China;* [e]*Information Hub, Hong Kong University of Science and Technology (Guangzhou), Guangzhou, China*

Contact by Email: juandu@ust.hk

# Tensor-based process control and monitoring for semiconductor manufacturing with unstable disturbances

With the development and popularity of sensors installed in manufacturing systems, complex data are collected during manufacturing processes, which brings challenges for traditional process control methods. This paper proposes a novel process control and monitoring method for the complex structure of high-dimensional image-based overlay errors (modeled in tensor form), which are collected in semiconductor manufacturing processes. The proposed method aims to reduce overlay errors using limited control recipes. We first build a high-dimensional process model and propose different tensor-on-vector regression algorithms to estimate parameters in the model to alleviate the curse of dimensionality. Then, based on the estimate of tensor parameters, the exponentially weighted moving average (EWMA) controller for tensor data is designed whose stability is theoretically guaranteed. Considering the fact that low-dimensional control recipes cannot compensate for all high-dimensional disturbances on the image, control residuals are monitored to prevent significant drifts of uncontrollable high-dimensional disturbances. Through extensive simulations and real case studies, the performances of parameter estimation algorithms and the EWMA controller in tensor space are evaluated. Compared with existing image-based feedback controllers, the superiority of our method is verified especially when disturbances are not stable.

## 1. Introduction

Process control and monitoring based on sensor data are important in advanced manufacturing such as semiconductor manufacturing, which aims to identify and reduce process variations caused by various disturbances during manufacturing processes. With recent achievements in sensing technology, extensive data with complex structures and high dimensions (such as multi-profile data, and image data) with heterogeneous resources can be collected in manufacturing systems (Xiang et al., 2021; Shi, 2023). For example, in semiconductor manufacturing, lithography as one of the most important processes aims to minimize the overlay error (i.e., misalignment) between adjacent material layers, which is considered as the quality variable in the manufacturing process and significantly affects the yield of final chips. Overlay errors in the lithography process are presented through wafer images. As shown in Figure 1(a), by measuring and marking a finite number of locations on a wafer, an image-based overlay error is obtained (Wang et al., 2021). At each measurement point, the directed overlay error is presented by an arrow, its length and direction signify the error magnitude and orientation respectively. Figure 1(c) quantifies the overlay error at each measurement point by projecting it into two directions (i.e., horizontal and vertical). Therefore, if the total number of measurement points located on wafers is $Q_1 \times Q_2$, the image-based overlay errors in Figure 1(a) can be described by structural tensor with dimension $Q_1 \times Q_2 \times 2$ in Figure 1(d).

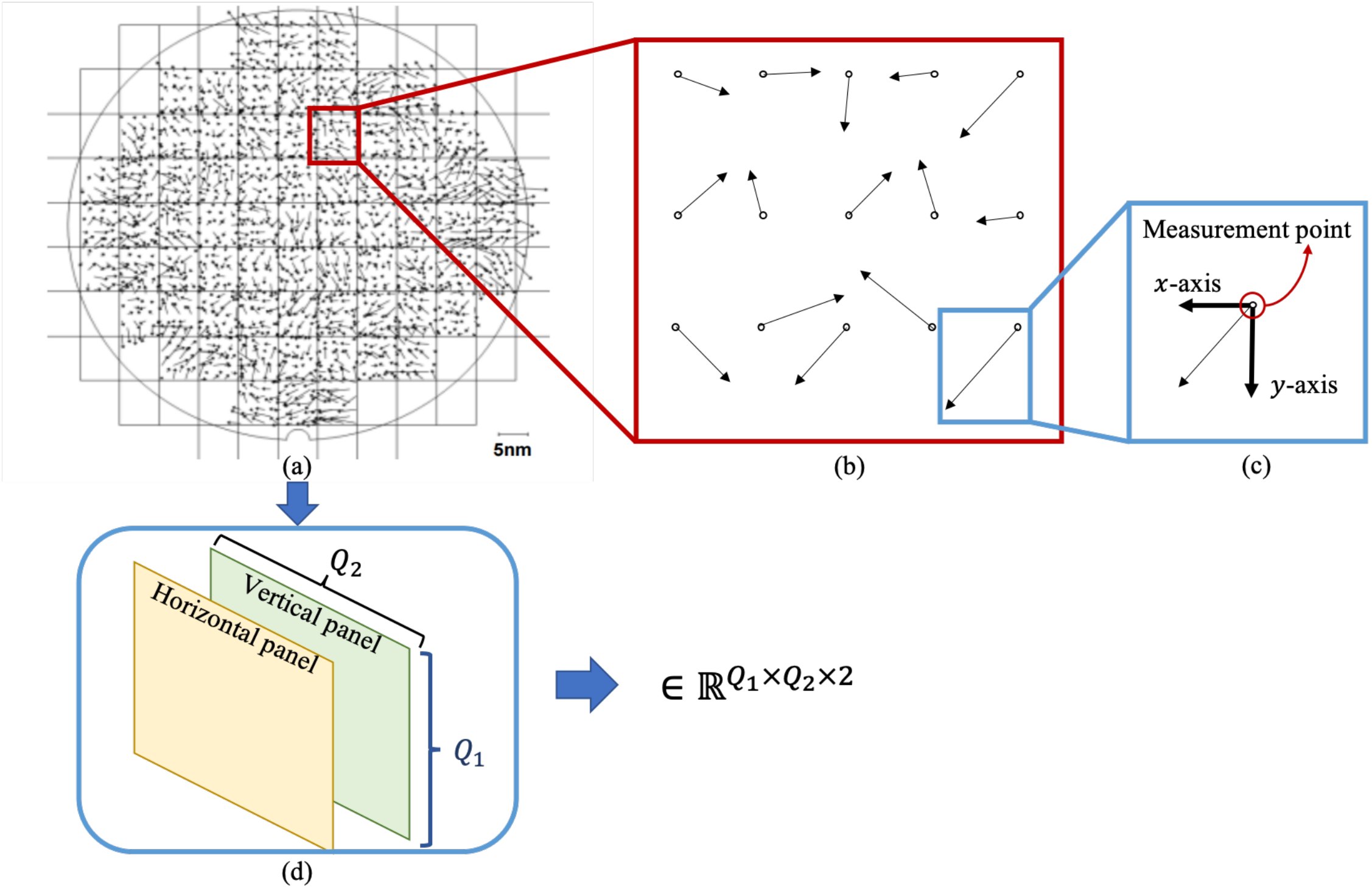

Figure 1. Illustration of overlay errors on a wafer

To reduce overlay errors during manufacturing processes, control recipes such as positions/orientations of wafers and the lens height can be adjusted to compensate for overlay errors (Zhang et al., 2023). However, the dimension of control recipes is limited and much less than that of overlay errors. It is difficult to design control recipes that completely compensate for all overlay errors with high dimensions, which brings challenges for traditional feedback control schemes in semiconductor manufacturing. Therefore, it is necessary to derive a novel control methodology with a tensor-on-vector process model to describe the relationship between high-dimensional quality data and limited control recipes.

Besides control recipes, unavoidable disturbances in manufacturing environments also significantly affect the system outputs and induce variations during manufacturing processes. In literature, disturbances are usually assumed to have the same dimensions as system outputs, which are usually defined as univariate or multivariate variables (Liu et al., 2018). Following this assumption, in lithography processes with image-based system outputs, overall disturbances also have the same dimensions as output overlay errors. Different from traditional process control problems with low dimensions, disturbance in lithography processes mainly sources from the following two main categories based on whether they can be effectively compensated by control recipes (Armitage and Kirk, 1988; Zhong et al., 2021).

The first type of disturbance can be interpreted by domain knowledge, which are attributed to known root causes. In lithography processes, different root causes can result in specific misalignment patterns. For example, local bumps are usually from chucking or lens distortion (Armitage and Kirk, 1988). If root causes are accurately verified, the corresponding control can be designed accordingly to compensate for the disturbance. In summary, this type of disturbance is low-dimensional and corresponds to certain root causes or control recipes.

The second type of disturbance sources from other random factors in the manufacturing environments such as dust. This type of disturbance exhibits similarly high dimensions as the image-based quality data and cannot be well-compensated through limited control adjustment. To prevent significant drifts in this type of disturbance, it is necessary to propose corresponding monitoring schemes following control implementation.

The two types of disturbances with different dimensions bring challenges to existing control schemes (Shkoruta et al., 2022), which are designed with a fixed dimension of control recipes (Ingolfsson and Sachs, 1993; Good and Qin, 2006). In summary, for manufacturing processes involving high-dimensional quality data, a novel process control scheme is necessary to tackle the following challenges:

- Complex quality data with high dimensions. Disturbances and quality variables involve high-dimensional data with complex structures, the corresponding parameters in the process model also exhibit complex structures with high dimensions that may be larger than the number of observations. Therefore, complex data bring challenges for existing parameter estimation and control optimization.
- Two types of hidden disturbances. Disturbances from different sources can induce overlay errors, but they cannot be directly measured and quantified by sensors. It is challenging to estimate identifiable or accurate parameters in the process model, thereby affecting the accuracy of control optimization.
- Limited control capability. As the second type of disturbance cannot be efficiently compensated by the limited number of control recipes, identifying and monitoring this type of disturbance based on real-time overlay errors is also a critical challenge.

To overcome these three challenges, we build a tensor-on-vector process model to describe the input-output relationship in manufacturing processes considering two types of disturbances. Figure 2 illustrates the framework of the proposed methodology, which consists of two phases. In phase I, the high-dimensional parameters in the process model are estimated by offline historical data. Meanwhile, monitoring statistics based on offline control residuals are also designed. In phase II, based on the parameter estimators, online control optimization and monitoring schemes are proposed.

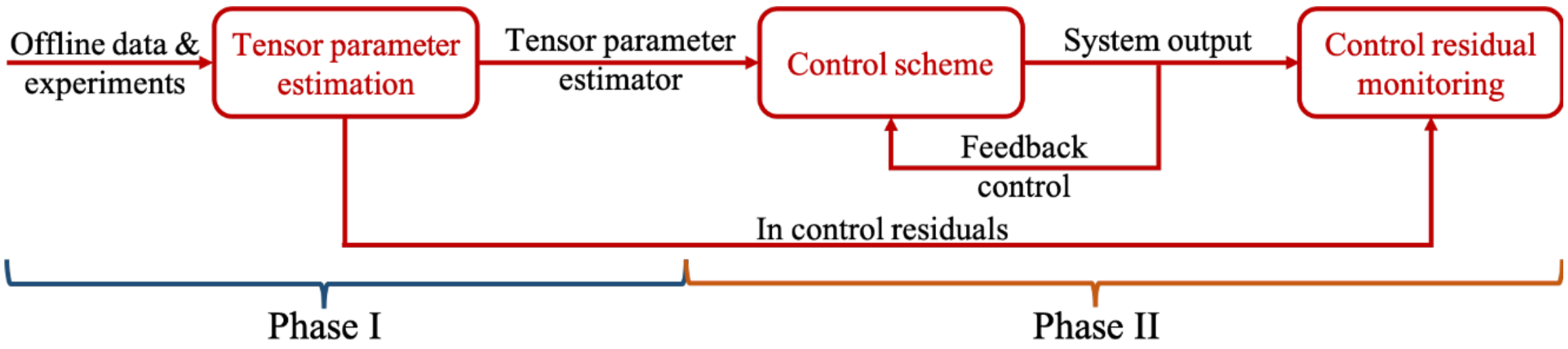

Figure 2. Tensor-based process control and monitoring

The remainder of this paper is organized as follows. Section 2 reviews related literature regarding process control and monitoring in tensor space. Section 3 proposes the tensor-based control methodology for tensor parameter estimation, control optimization, and residual monitoring, respectively. The corresponding algorithms and theorems are also presented. Section 4 presents extensive simulations to verify the performance of different parameter estimation algorithms and control schemes. Section 5 further compares the performances of the proposed controller with an existing image-based feedback controller through a case study of lithography processes. The characteristics of different tensor-based monitoring methods are also discussed. Finally, the conclusions and future research are summarized in Section 6.

## 2. Literature review

Process control and monitoring are important research topics in many complex manufacturing systems, such as additive manufacturing (Liang et al., 2004), steel industries (Kano and Nakagawa, 2008), and semiconductor manufacturing (Liu et al., 2018). Taking semiconductor manufacturing as an example, extensive run-to-run (R2R) control and monitoring methodologies are discussed based on univariate or multivariate system inputs and outputs. We will present the research in two aspects related to control and monitoring respectively.

For extensive R2R control schemes, Ingolfsson and Sachs (1993) proposed the EWMA controller for single input-singe output (SISO) systems and analyzed its stability. Good and Qin (2006) extended the EWMA controller into the multiple input-multiple output (MIMO) systems and found its stability is still guaranteed. Chen and Chuang (2010) applied the MIMO EWMA control schemes to the chemical-mechanical polishing process. Apley and Lee (2010) verified the efficiency of the EWMA scheme in time series forecasting, statistical process control, and automatic feedback control. Therefore, the EWMA control scheme holds significant potential to tackle more complex data.

For different monitoring methodologies in semiconductor manufacturing processes, Ge and Song (2010) introduced an adaptive sub-statistical principal component analysis (PCA)-based monitoring method to deal with non-Gaussian process data with a limited number of batches. Different from PCA-based monitoring methods, Yu (2012) proposed a new feature extraction method by the Gaussian mixture model and Bayesian method. Zou et al. (2015) focused on high-dimensional streaming data and designed a new online monitoring method based on a powerful goodness-of-fit test of the local cumulative sum statistics from each data stream. Gentner et al. (2022) systematically summarized different process monitoring methods applied in semiconductor manufacturing. In summary, extensive process control and monitoring methods focus on univariate or multivariate data, but image-based methodologies are limited.

Tensor analysis is an efficient method to deal with data with complex structures and high dimensions, such as image data. In literature, two mainstream research related to our work are tensor-based regression and monitoring methods. Tensor regression methods are based on the assumption that the high-dimensional coefficients have an appropriate low-dimensional structure. For example, Zhou et al. (2013) proposed a variable-on-tensor regression model and applied it to neuroimaging data analysis. Lock (2018) extended the linear regression model into tensor space, which is formulated as a tensor-on-tensor regression model. Then the parameters were estimated and the corresponding identifiable properties were guaranteed. Gahrooei et al. (2021) introduced a multiple tensor-on-tensor regression approach for multiple input tensors with different orders to deal with heterogeneous sources of data. Moreover, other factors on tensor regression are also analyzed, such as the interactions of input tensor (Miao et al., 2021), missing values (Wang et al., 2022), partial observations (Zhang et al., 2023), and outliers (Lee et al, 2024).

Tensor-based monitoring methods are widely applied in image data for anomaly detection and fault diagnosis. For example, Hu and Yuan (2009) designed a batch-based monitoring method using tensor locality preserving projection and applied it to industrial fermentation processes. Yan et al. (2014) introduced different image-based process monitoring methods based on low-rank tensor decomposition such as unfold PCA, multilinear PCA, uncorrelated PCA, tensor rank-one decomposition, and applied these methods in steel tube manufacturing to detect process changes. Khanzadeh et al. (2018) focused on the complex spatial interdependence and low sign-to-noise ratios in the image data and proposed an online dual control charting system to detect the changes in the features and residuals. Luo et al. (2020) proposed an adaptive sequential fault diagnosis method based on a tensor factorization layer merged with deep neural networks.

However, only a few works focus on image-based process control. To our knowledge, in semiconductor manufacturing processes, Zhong et al. (2023) designed an image-based feedback control scheme considering the spatio-temporal correlations by tensor analysis. Zhang et al. (2023) focused on image-based control considering partial observations. There is limited research that focuses on high-dimensional unstable disturbances and combines automatic process control with statistical process control in tensor space for image-based quality analytics. Therefore, to deal with two types of disturbances for image-based process control, we propose a novel methodology to estimate the high-dimensional parameters in the process model, optimize control recipes at each run during the manufacturing process, and monitor the control residuals at the same time.

## 3. Methodology

In this section, a tensor-based methodology for process control and monitoring is introduced to reduce overlay errors evaluated by image data. Specifically, we first introduce the formulations of the tensor-based process control and monitoring problem in Section 3.1. In Section 3.2, different estimation methods of high-dimensional parameters in the process model are proposed to obtain identifiable estimators. In Section 3.3, based on parameter estimators in Section 3.2, the exponentially weighted moving average (EWMA) scheme is extended in tensor space for control optimization, and the theoretical stability region is analyzed. Finally, different tensor-based control charts are proposed based on control residuals for monitoring the second type of disturbance in Section 3.4.

### *3.1. Problem description and formulation*

Through this paper, we use lowercase letters to denote a scalar (such as $a$); lowercase boldface letters to denote a vector (such as $\boldsymbol{a}$); uppercase boldface letters to denote a matrix (such as $\boldsymbol{A}$); and calligraphic letters to denote a tensor (such as $\mathcal{A}$). Based on these notations, we consider the tensor-based process modeling for lithography processes as follows.

Consider a batch-based R2R manufacturing process with a total number of runs $T$, i.e., a sequence of wafers can be collected from run 1 to $T$ in a production cycle. At each run $t \in \{1,2,\dots,T\}$, an image-based quality variable, as illustrated in Figure 1, is collected and denoted as $\mathcal{Y}_t \in \mathbb{R}^{Q_1 \times Q_2 \times Q_3}$. $Q_1$ and $Q_2$ represent dimensions of measurement points on the wafer, and $Q_3 = 2$ in this work corresponds to the overlay errors measured along the horizontal ($x$-axis) and vertical directions ($y$-axis), respectively, at each point. As two types of disturbances both significantly affect system outputs, we define them respectively as follows.

The first type of disturbance sources from known root causes, which is defined as $\boldsymbol{d}_t \in \mathbb{R}^m$ at run $t$, and $m$ denotes the number of root causes. Specifically, for each root cause, the corresponding input control variable is defined as $\boldsymbol{u}_t \in \mathbb{R}^m$, which can be designed to reduce the overlay errors by adjusting the parameter knobs. The second type of disturbance denotes other factors in the manufacturing environments without known root causes, which are defined as $\mathcal{E}_t \in \mathbb{R}^{Q_1 \times Q_2 \times Q_3}$. In summary, the process model is:

$$\mathcal{Y}_t = (\boldsymbol{u}_t + \boldsymbol{d}_t) * \mathcal{B} + \mathcal{E}_t, \tag{1}$$

where $\mathcal{B} \in \mathbb{R}^{m \times Q_1 \times Q_2 \times Q_3}$ is the tensor parameter.

Before optimizing the control recipes to minimize overlay errors, finding an accurate estimator for high-dimensional tensor parameter $\mathcal{B}$ is important. However, during manufacturing processes, only input control recipes $\boldsymbol{u}_t$ and output images $\mathcal{Y}_t$ can be observed. As two types of disturbances $\boldsymbol{d}_t$ and $\mathcal{E}_t$ are both unobservable, obtaining an identifiable estimator for $\mathcal{B}$ is challenging. Therefore, in Section 3.2, we discuss methodologies for tensor parameter estimation based on these two types of disturbances.

### *3.2. Parameter estimation*

Before online control and monitoring can be implemented in phase II, it is crucial to estimate the tensor parameters in the process model in Equation (1), and the estimation is typically achieved by using historical or experimental offline data. Different from process control problems in low-dimensional spaces, the accuracy of estimating tensor parameters plays a more crucial role in process control problems, as high-dimensional tensor parameters can significantly amplify the effects of control recipes. Therefore, in this subsection, we aim to minimize the difference between the parameter estimator $\widehat{\mathcal{B}}$ and the ground truth $\mathcal{B}$.

According to the discussion of the process model in Section 3.1, disturbances $\boldsymbol{d}_t$ and $\mathcal{E}_t$ are unobserved, and only input control recipes $\boldsymbol{u}_t$ and output images $\mathcal{Y}_t$ at each run $t$ can be collected. So we discuss different algorithms for parameter estimation based on offline data $\boldsymbol{u}_t$ and $\mathcal{Y}_t$.

*A. Basic least square estimation in tensor space*

Least square (LS) is a fundamental method for parameter estimation, which can also be extended in tensor space for image-based quality data. One of the most intractable challenges is estimating a large number of elements in the tensor parameter. To deal with this curse of dimensionality, in literature such as Gahrooei et al. (2021) and Zhong et al., (2023), parameter $\mathcal{B}$ can be supposed to lie in lower dimensional spaces and expanded using a set of basis matrices by tensor product, i.e., the parameter $\mathcal{B}$ can be expressed as

$$\mathcal{B} = \mathcal{C}_{\mathcal{B}} \times_2 \boldsymbol{V}_1 \times_3 \boldsymbol{V}_2 \times_4 \boldsymbol{V}_3, \tag{2}$$

where $\mathcal{C}_{\mathcal{B}} \in R^{m \times P_1 \times P_2 \times P_3}$ is the core tensor of $\mathcal{B}$ with much lower dimensions ($P_i \ll Q_i$, $i = 1,2,3$). $\boldsymbol{V}_i$ ($i = 1,2,3$) are basis matrices with dimensions as $P_i \times Q_i$.

As $\boldsymbol{d}_t$ and $\mathcal{E}_t$ are hidden variables, we ignore their effects and employ the existing tensor regression methods based on the observed data $\boldsymbol{u}_t$ and $\mathcal{Y}_t$ in this part. Yan et al. (2019) and Gahrooei et al. (2021) have considered the case where disturbances are independent and identically distributed and extended the LS method into tensor space for parameter estimation. We define this estimation method as tensor-on-vector regression by LS. Specifically, suppose there are $N$ historical production cycles in historical offline dataset, and $T$ runs totally in each production cycle. We use $\mathcal{Y}$ to denote the set of total system output images from run 1 to $T$ in $N$ historical offline cycles. Therefore, we have $\mathcal{Y} \in \mathbb{R}^{n \times Q_1 \times Q_2 \times Q_3}$, where $n = N * T$ denotes the number of observations in the offline dataset. Meanwhile, $\boldsymbol{U} \in \mathbb{R}^{n \times m}$ is the set of the corresponding input control recipes. Then, the core tensor ($\mathcal{C}_{\mathcal{B}}$) and basis matrices ($V_i$, $i = 1,2,3$) can be solved by the following optimization problems:

$$\min_{\mathcal{C}_{\mathcal{B}}, V_1, V_2, V_3,} \left\| \boldsymbol{Y}_{(1)} - \boldsymbol{U} * \boldsymbol{B}_{(1)} \right\|_F^2$$

$$\text{s.t. } \mathcal{B} = \mathcal{C}_{\mathcal{B}} \times_2 \boldsymbol{V}_1 \times_3 \boldsymbol{V}_2 \times_4 \boldsymbol{V}_3$$

$$\boldsymbol{V}_i^{\boldsymbol{T}} \boldsymbol{V}_i = \boldsymbol{I}_{P_i} \; (i = 1,2,3), \tag{3}$$

where $\boldsymbol{Y}_{(i)}$ and $\boldsymbol{B}_{(i)}$ are the mode-$i$ unfolding of tensors $\mathcal{Y}$ and $\mathcal{B}$ respectively, and $\boldsymbol{I}_{P_i}$ is a $P_i \times P_i$ identify matrix.

Following the methodology of tensor-on-vector regression by LS, we can combine alternating least square (ALS) and block coordinate descent (BCD) methods to solve $\mathcal{C}_{\mathcal{B}}$ and $\boldsymbol{V}_i$ ($i = 1,2,3$) iteratively. Then, the tensor parameter estimator is obtained according to Equation (2). Specifically, according to Gahrooei et al. (2021), there are two steps to solve $\mathcal{C}_{\mathcal{B}}$ and $\boldsymbol{V}_i$ ($i = 1,2,3$), respectively.

**Step 1**: Given the initial values of $\boldsymbol{V}_i$, the core tensor $\mathcal{C}_{\mathcal{B}}$ can be solved by LS methods as

$$\mathcal{C}_{\mathcal{B}} = \mathcal{Y} \times_1 (\boldsymbol{X}^T\boldsymbol{X})^{-1}X^T \times_2 (\boldsymbol{V}_1^T\boldsymbol{V}_1)^{-1}\boldsymbol{V}_1^T \times_3 (\boldsymbol{V}_2^T\boldsymbol{V}_2)^{-1}\boldsymbol{V}_2^T \times_4 (\boldsymbol{V}_3^T\boldsymbol{V}_3)^{-1}\boldsymbol{V}_3^T, \quad (4)$$

where $\boldsymbol{X} = \boldsymbol{U} + \boldsymbol{D}$, and $\boldsymbol{D}$ is the set of estimators for $\boldsymbol{d}_t$. As disturbances cannot be directly observed, we initialize $\boldsymbol{D}$ as $\boldsymbol{0}$ and update its value according to parameter estimators with each iteration.

**Step 2**: Given the core tensor $\mathcal{C}_{\mathcal{B}}$, solving the basis matrices in the optimization problem in Equation (3) can be transformed into an orthogonal Procrustes problem with the solution:

$$\boldsymbol{V}_i = \boldsymbol{R}_i\boldsymbol{W}_i^T, \quad (5)$$

where $\boldsymbol{R}_i$ and $\boldsymbol{W}_i^T$ are obtained from the singular value decomposition of $\boldsymbol{Y}_{(i)}\boldsymbol{S}_i^T$, and $\boldsymbol{S}_i = \boldsymbol{C}_{\mathcal{B}(i)} \prod_{j \neq i} \boldsymbol{V}_j$, and $\boldsymbol{C}_{\mathcal{B}(i)}$ is the mode-$i$ unfolding of tensors $\mathcal{C}_{\mathcal{B}}$.

By calculating $\mathcal{C}_{\mathcal{B}}$ and $\boldsymbol{V}_i$ iteratively until convergence, we get the estimator $\widehat{\mathcal{B}} = \mathcal{C}_{\mathcal{B}} \times_2 \boldsymbol{V}_1 \times_3 \boldsymbol{V}_2 \times_4 \boldsymbol{V}_3$. Finally, by substituting the tensor parameter estimator $\mathcal{B}$ into the process model in Equation (1), we have the estimation value of two types of disturbances. Algorithm 1 presents the details of the tensor-on-vector regression by LS.

Unfortunately, the parameter estimator $\widehat{\mathcal{B}}$ calculated by Algorithm 1 is usually biased. The reason is that we can only collect $\boldsymbol{u}_t$ and $\mathcal{Y}_t$, it is insufficient to estimate an identifiable estimator for $\mathcal{B}$ based on the process model in Equation (1) (Xing et al., 2021). As $\mathcal{E}_t$ and $\boldsymbol{d}_t$ are both hidden variables, and $\mathcal{E}_t$ denotes high-dimensional disturbance that cannot be compensated by $\boldsymbol{u}_t$, it is reasonable to assume that they are independent. Therefore, the bias of $\widehat{\mathcal{B}}$ primarily attributes to $\boldsymbol{d}_t$. The following proposition and corollary analyze the performance of $\widehat{\mathcal{B}}$ in Algorithm 1 based on correlations between $\boldsymbol{d}_t$ and $\boldsymbol{u}_t$.

**Algorithm 1: Algorithm of tensor-on-vector regression by LS**

[input] $\boldsymbol{U}$ and $\mathcal{Y}$

[Initialization] $\boldsymbol{D} = \mathbf{0}, \widehat{\boldsymbol{V}}_1, \widehat{\boldsymbol{V}}_2, \widehat{\boldsymbol{V}}_3$

**Replicate:**

**Replicate:**

$$\mathcal{C}_{\mathcal{B}} = \mathcal{Y} \times_1 \left((\boldsymbol{U} + \boldsymbol{D})^T(\boldsymbol{U} + \boldsymbol{D})\right)^{-1}(\boldsymbol{U} + \boldsymbol{D})^T \times_2 (\widehat{\boldsymbol{V}}_1^T\widehat{\boldsymbol{V}}_1)^{-1}\widehat{\boldsymbol{V}}_1^T \times_3 (\widehat{\boldsymbol{V}}_2^T\widehat{\boldsymbol{V}}_2)^{-1}\widehat{\boldsymbol{V}}_2^T \times_4 (\widehat{\boldsymbol{V}}_3^T\widehat{\boldsymbol{V}}_3)^{-1}\widehat{\boldsymbol{V}}_3^T$$

$\widehat{\boldsymbol{V}}_i = \boldsymbol{R}_i\boldsymbol{W}_i^T$, $\boldsymbol{R}$ and $\boldsymbol{W}^T$ are obtained from the singular value decomposition of $\boldsymbol{Y}_{(i)}\boldsymbol{S}_i^T$, where $\boldsymbol{S}_i = \boldsymbol{C}_{\mathcal{B}(i)} \prod_{j\neq i} \widehat{\boldsymbol{V}}_j$.

$\widehat{\mathcal{B}} = \mathcal{C}_{\mathcal{B}} \times_2 \widehat{\boldsymbol{V}}_1 \times_3 \widehat{\boldsymbol{V}}_2 \times_4 \widehat{\boldsymbol{V}}_3$

$\mathcal{E} = \mathcal{Y} - (\boldsymbol{U} + \boldsymbol{D}) * \widehat{\mathcal{B}}$, and $e = \left\|\boldsymbol{\mathcal{E}}_{(1)}\right\|_F$

**Until**: $e$ converges

Update the disturbances by $\boldsymbol{D} = \boldsymbol{C}_{\mathcal{B}(1)}^{-1} * (\mathcal{Y} \times_2 \widehat{\boldsymbol{V}}_1^T \times_3 \boldsymbol{V}_2^T \ldots \times_4 \boldsymbol{V}_3^T - \boldsymbol{U} * \mathcal{C}_{\mathcal{B}})$

**Until**: $\widehat{\mathcal{B}}$ converges

[Output] $\widehat{\mathcal{B}}$ and $\mathcal{E} = \mathcal{Y} - (\boldsymbol{U} + \boldsymbol{D}) * \widehat{\mathcal{B}}$.

---

**Proposition 1:** *The parameter estimator $\widehat{\mathcal{B}}$ is not identifiable, which depends on linear correlations between $\boldsymbol{d}_t$ and $\boldsymbol{u}_t$. If $\boldsymbol{d}_t$ can be formulated as*

$$\boldsymbol{d}_t = \boldsymbol{A}^T\boldsymbol{u}_t + \boldsymbol{W}_t. \tag{6}$$

*We have $E(\widehat{\mathcal{B}}) = (\boldsymbol{I}_m + \boldsymbol{A}) * \mathcal{B}$, where $\boldsymbol{I}_m$* is an identify matrix with dimension $m \times m$. The proof is provided in Appendix A.

**Corollary 1:** *When $\boldsymbol{d}_t$ is independent of $\boldsymbol{u}_t$, the estimator $\widehat{\mathcal{B}}$ obtained by Algorithm 1 is unbiased.*

In practical semiconductor manufacturing processes, disturbances ($\boldsymbol{d}_t$ and $\mathcal{E}_t$) at each run source from manufacturing systems or environments, which are usually assumed to be independent of control recipes $\boldsymbol{u}_t$, especially when disturbances are weakly autocorrelated (Del Castillo and Hurwitz, 1997; Liu et al., 2018). However, if the first type of disturbance exhibits significant autocorrelations in offline data, the condition in Corollary 1 will not hold. In this case, $\boldsymbol{u}_t$ inferred by $\mathcal{Y}_{t-1}$ (or $\boldsymbol{d}_{t-1}$) may correlate with $\boldsymbol{d}_t$, and Algorithm 1 ignores this correlation, thereby leading to bias of estimator $\widehat{\mathcal{B}}$.

Based on Proposition 1, although we cannot estimate an unidentifiable and unbiased parameter estimator for the process model in Equation (1), it is necessary to consider additional conditions and improve the performance of Algorithm 1, especially when correlations between $\boldsymbol{u}_t$ and $\boldsymbol{d}_t$ are large. Therefore, we specify other familiar cases on parameter $\mathcal{B}$ and propose corresponding estimation methods for a more accurate tensor estimator as follows.

*B. Extension estimation algorithm using sparse learning*

In lithography processes, different root causes in the manufacturing process are typically reflected by their respective components or parameters, which can be independent. Moreover, a certain type of overlay error is attributed to a limited number of root causes, which implies that the parameter tensor to describe the relationship between overlay errors and root causes is row sparsity. In summary, we have the following two assumptions as follows.

**Assumption 1:** *Different root causes for the first type of disturbance are independent.*

**Assumption 2:** *The core of parameter tensor $\mathcal{B}$ satisfies the condition that its Mode-1 unfolding matrix $\boldsymbol{C}_{\mathcal{B}(1)}$ is row-sparse.*

Considering that disturbances are not independent of control actions, following Equation (6) in Proposition 1, we divide the disturbance $\boldsymbol{d}_t$ into two parts: $\boldsymbol{A}^T\boldsymbol{u}_t$ $(\boldsymbol{A} \neq \boldsymbol{0})$ and $\boldsymbol{W}_t$. To overcome the limitation of Algorithm 1 when the correlation parameter $\boldsymbol{A}$ is large, we improve LS estimation in Algorithm 1 by fully considering the correlations between $\boldsymbol{d}_t$ and $\boldsymbol{u}_t$. If Assumptions 1 and 2 hold in manufacturing processes, Proposition 2 presents the details of estimating the tensor parameter by combining group lasso and ridge problem (GLRP).

**Proposition 2:** *Based on Assumptions 1 and 2, we can obtain the core tensor of the parameter estimator $\mathcal{C}_{\mathcal{B}}$ by the optimization problem:*

$$\text{(GLRP)} \qquad \min_{\boldsymbol{C}_{\mathcal{B}(1)},\boldsymbol{A}} \frac{1}{n}\left\|\boldsymbol{C}_{\mathcal{Y}(1)} - \boldsymbol{U} * (\boldsymbol{I} + \boldsymbol{A})\boldsymbol{C}_{\mathcal{B}(1)}\right\|_F^2 + \lambda_1\left\|\boldsymbol{C}_{\mathcal{B}(1)}\right\|_{l_1/l_2} + \lambda_2\|\boldsymbol{A}\|_F^2, \qquad (7)$$

*where $\lambda_1$ and $\lambda_2$ are tuning parameters, $\boldsymbol{C}_{\mathcal{Y}(1)}$ is the mode-1 unfolding of $\mathcal{C}_{\mathcal{Y}}$, and $\mathcal{C}_{\mathcal{Y}} = \mathcal{Y} \times_2 \boldsymbol{V}_1^T \times_3 \boldsymbol{V}_2^T \times_4 \boldsymbol{V}_3^T$. Moreover, $\|\cdot\|_{l_1/l_2}$ is defined on a matrix (whose dimension is $d_1 \times d_2$) with formulation as $\|\boldsymbol{M}\|_{l_1/l_2} = \sum_{j=1}^{d_1}\left\|\boldsymbol{M}_{j\cdot}\right\|_{l_2}$, where $\boldsymbol{M}_{j\cdot}$ denotes the jth row of matrix **M**.*

According to the BCD method in Chernozhukov et al. (2017) and Bing et al. (2023), the solution of GLRP in Equation (7) can be obtained by solving the following two sub-problems, i.e., group lasso in Equation (8) and ridge regression in Equation (9) iteratively until the solutions convergent.

$$\widehat{\boldsymbol{C}}_{\mathcal{B}(1)} = \arg\min_{\boldsymbol{C}_{\mathcal{B}}} \frac{1}{n}\left\|\boldsymbol{C}_{\mathcal{Y}(1)} - \boldsymbol{U}\big(\boldsymbol{I} + \widehat{\boldsymbol{A}}\big)\boldsymbol{C}_{\mathcal{B}(1)}\right\|_F^2 + \lambda_1\left\|\boldsymbol{C}_{\mathcal{B}(1)}\right\|_{l_1/l_2}, \qquad (8)$$

$$\widehat{\boldsymbol{A}} = \arg\min_{\boldsymbol{A}} \frac{1}{n}\left\|\boldsymbol{C}_{\mathcal{Y}(1)} - \boldsymbol{U}(\boldsymbol{I} + \boldsymbol{A})\widehat{\boldsymbol{C}}_{\mathcal{B}(1)}\right\|_F^2 + \lambda_2\|\boldsymbol{A}\|_F^2. \qquad (9)$$

After the convergence of $\widehat{\boldsymbol{C}}_{\mathcal{B}(1)}$ and $\widehat{\boldsymbol{A}}$, we obtain the solution to the GLRP optimization problem. By replacing the LS step in Algorithm 1 with GLRP, an improved algorithm for a more accurate estimator of the tensor parameter is presented in Algorithm 2.

**Algorithm 2: Algorithm of tensor-on-vector regression by GLRP**

[input] $\boldsymbol{U}$ and $\mathcal{Y}$

[Initialization] $\boldsymbol{D} = \boldsymbol{0}, \widehat{\boldsymbol{V}}_1, \widehat{\boldsymbol{V}}_2, \widehat{\boldsymbol{V}}_3$

**Replicate:**

**Replicate:**

$$\left(\widehat{\boldsymbol{C}}_{\mathcal{B}(1)}, \widehat{\boldsymbol{A}}\right) = \arg\min_{\boldsymbol{C}_{\mathcal{B}(1)}, \boldsymbol{A}} \frac{1}{n} \left\|\boldsymbol{C}_{\mathcal{Y}(1)} - \boldsymbol{U} * (\boldsymbol{I} + \boldsymbol{A})\boldsymbol{C}_{\mathcal{B}(1)}\right\|_F^2 + \lambda_1 \left\|\boldsymbol{C}_{\mathcal{B}(1)}\right\|_{l_1/l_2} + \lambda_2 \|\boldsymbol{A}\|_F^2$$

$\widehat{\boldsymbol{V}}_i = \boldsymbol{R}_i \boldsymbol{W}_i^T$, $\boldsymbol{R}$ and $\boldsymbol{W}^T$ are obtained from the singular value decomposition of $\boldsymbol{Y}_{(i)} \boldsymbol{S}_i^T$, where $\boldsymbol{S}_i = \widehat{\boldsymbol{C}}_{\mathcal{B}(i)} \prod_{j \neq i} \boldsymbol{V}_j$.

$\widehat{\mathcal{B}} = \hat{\mathcal{C}}_{\mathcal{B}} \times_2 \widehat{\boldsymbol{V}}_1 \times_3 \widehat{\boldsymbol{V}}_2 \times_4 \widehat{\boldsymbol{V}}_3$

$\mathcal{E} = \mathcal{Y} - (\boldsymbol{U} + \boldsymbol{D}) * \widehat{\mathcal{B}}$, and $e = \left\|\boldsymbol{\mathcal{E}}_{(1)}\right\|_F$

**Until:** $e$ converges

Update the disturbances by $\boldsymbol{D} = \widehat{\boldsymbol{C}}_{\mathcal{B}(1)}^{-1} * \left(\mathcal{Y} \times_2 \widehat{\boldsymbol{V}}_1^T \times_3 \widehat{\boldsymbol{V}}_2^T \dots \times_4 \widehat{\boldsymbol{V}}_3^T - \boldsymbol{U} * \hat{\mathcal{C}}_{\mathcal{B}}\right)$

**Until**: $\widehat{\mathcal{B}}$ converges

[Output] $\widehat{\mathcal{B}}$ and $\mathcal{E} = \mathcal{Y} - (\boldsymbol{U} + \boldsymbol{D}) * \widehat{\mathcal{B}}$.

### *3.3. Online control scheme*

Based on a parameter estimator $\widehat{\mathcal{B}}$ in the process model, a corresponding control scheme can be subsequently designed. In literature, one of the prevalent methods in R2R control schemes is the EWMA controller, which keeps the capability of being extended into tensor space. Similar to single input-single output (SISO) or multiple input-multiple output (MIMO) process control problems, we can also derive the stability region of the EWMA controller in tensor space.

Following the R2R control objective to keep the quality variable close to the desired level, we tend to keep the overlay error close to $\mathbf{0} \in \mathbb{R}^{Q_1 \times Q_2 \times Q_3}$ in this work. Based on the EWMA control scheme, at each run $t$, we have the following optimization problem to obtain the control recipes.

$$
\begin{aligned}
&\min_{\boldsymbol{u}_t} \sum_{t=1}^{T} \left\| \mathrm{vec}(\hat{\mathcal{Y}}_t) \right\|_F \\
\text{s.t.} \quad & \hat{\mathcal{C}}_{\mathcal{B}} = \hat{\mathcal{B}} \times_2 \hat{\boldsymbol{V}}_1^T \times_3 \hat{\boldsymbol{V}}_2^T \times_4 \hat{\boldsymbol{V}}_3^T \\
& \hat{\mathcal{Y}}_t = (\boldsymbol{u}_t * \hat{\mathcal{C}}_{\mathcal{B}} + \hat{d}_t) \times_1 \hat{\boldsymbol{V}}_1 \times_2 \hat{\boldsymbol{V}}_2 \times_3 \hat{\boldsymbol{V}}_3 \\
& \mathcal{C}_{\mathcal{Y}_t} = \mathcal{Y}_t \times_1 \hat{\boldsymbol{V}}_1^T \times_2 \hat{\boldsymbol{V}}_2^T \times_3 \hat{\boldsymbol{V}}_3^T \\
& \hat{d}_{t+1} = \lambda\left(\mathcal{C}_{\mathcal{Y}_t} - \hat{\mathcal{C}}_{\mathcal{B}} \times_1 \boldsymbol{u}_t\right) + (1-\lambda)\hat{d}_t \\
& \boldsymbol{u}_{t+1} = -\left(\hat{\boldsymbol{C}}_{\mathcal{B}(1)}^T \hat{\boldsymbol{C}}_{\mathcal{B}(1)}\right)^{-1} \hat{\boldsymbol{C}}_{\mathcal{B}(1)}^T * \mathrm{vec}(\hat{d}_{t+1}),
\end{aligned} \tag{10}
$$

where $\mathrm{vec}(\hat{\mathcal{Y}}_t)$ denotes the vectorization of tensor $\hat{\mathcal{Y}}_t$, $\lambda \in (0,1)$ is the tuning parameter, and $\hat{d}_t \in \mathbb{R}^{P_1 \times P_2 \times P_3}$ is the estimated overlay error caused by disturbances before control implementation based on the EWMA scheme. Following the assumption that the high-dimensional image variable can be well interpreted in the core space with lower dimensions, we predict the first type of disturbance in the core space according to the EWMA scheme. The first three constraints denote the transformation of parameter, predicted and real system outputs in tensor and their core spaces, respectively. The fourth constraint implies the disturbance prediction in the next run, and the last constraint represents the formulation of control optimization. The stability of this control scheme is guaranteed in Theorem 1.

**Theorem 1**: *Let* $\boldsymbol{M} = \boldsymbol{I} - \boldsymbol{\Lambda}\boldsymbol{\xi}$ , *where* $\boldsymbol{\Lambda} = \lambda \boldsymbol{I}$ , *and* $\boldsymbol{\xi} = \left(\mathcal{B} \times_2 \hat{\boldsymbol{V}}_1^T \times_3 \hat{\boldsymbol{V}}_2^T \times_3 \dots \times_{d+1} \hat{\boldsymbol{V}}_d^T\right)_{(1)} \left(\hat{\boldsymbol{C}}_{\mathcal{B}(1)}^T \hat{\boldsymbol{C}}_{\mathcal{B}(1)}\right)^{-1} \hat{\boldsymbol{C}}_{\mathcal{B}(1)}^T$, $\hat{\boldsymbol{C}}_{\mathcal{B}(1)}$ *is mode-1 unfold for the core tensor of the estimated parameter. The stability condition of the EWMA controller* *in tensor space is that the absolute values of all eigenvalues of* $\boldsymbol{M}$ *are less than 1 (i.e.,* $|\text{eig}(\boldsymbol{M})| < 1$*). Specifically, if* $\text{eig}(\boldsymbol{M}) = a_j + b_j i$*, the stability condition is:* $\frac{2a_j}{a_j^2+b_j^2} > \lambda$*.*

The proof is provided in Appendix B.

In particular, if Algorithm 1 is chosen for tensor parameter estimation, the stability region can be specified in Corollary 2.

**Corollary 2:** *If Algorithm 1 is used to estimate tensor parameter* $\mathcal{B}$*, the stable condition of the EWMA controller is* $\text{eig}(\boldsymbol{A}) > \frac{\lambda-2}{2}$*, where* $\boldsymbol{A}$ *is the correlation parameter between disturbances and control recipes in Equation (6).*

The proof is provided in Appendix C.

In summary, similar to SISO and MIMO control systems, the EWMA control scheme in tensor space also has a stability region, which depends on the turning parameter ($\lambda$) and the accuracy of the tensor parameter estimator ($\hat{\mathcal{B}}$). We also verify the performance of the EWMA controller by numerical studies in Sections 4 and 5.

### *3.4. Residual monitoring*

During the batch-based lithograph processes, after applying control recipes at run $t$, the image-based quality variable $\mathcal{Y}_t$ is observed. As we discussed before, besides $\boldsymbol{d}_t$ sources from typical root causes, $\mathcal{E}_t$ from the manufacturing environments also influences $\mathcal{Y}_t$. Due to the limited dimensions, the control recipe is incapable of fully compensating for the high-dimensional disturbances $\mathcal{E}_t$, and we propose monitoring methods based on $\mathcal{Y}_t$ to identify whether a significant drift has occurred to $\mathcal{E}_t$ during manufacturing processes. If an out-of-control alarm occurs, which means that general control recipes cannot compensate for disturbances, this batch of products needs to be held up and the manufacturing system requires a reset to exclude disturbances.

However, the system output $\mathcal{Y}_t$ cannot directly reflect the second type of disturbance $\mathcal{E}_t$, due to the prediction errors of $\boldsymbol{d}_t$, which also result in the increment of $\mathcal{Y}_t$. Therefore, it is important to identify the effects of $\mathcal{E}_t$ at first. The fundamental difference between the two types of disturbances is whether they can be efficiently compensated by control recipes, so we use tensor projection on the control space to divide them. The portion of $\mathcal{Y}_t$ that can be projected into control space by basis matrices is considered as the prediction error of $\boldsymbol{d}_t$. Otherwise, it is induced by $\mathcal{E}_t$. Therefore, Equation (11) expresses the estimation of the $\mathcal{E}_t$, which is defined as the control residuals of the recipes.

$$\mathcal{R}_t = \mathcal{Y}_t - \mathcal{Y}_t \times_1 \left(\widehat{\boldsymbol{V}}_1^T \widehat{\boldsymbol{V}}_1\right) \times_2 \left(\widehat{\boldsymbol{V}}_2^T \widehat{\boldsymbol{V}}_2\right) \times_3 \left(\widehat{\boldsymbol{V}}_3^T \widehat{\boldsymbol{V}}_3\right). \quad (11)$$

The monitoring methodology in this subsection is primarily targeted at $\mathcal{R}_t$. Suppose that if the manufacturing batch proceeded without suspended, the control residuals are regarded as states of in-control. By selecting the complete offline manufacturing batches, we construct the statistics in phase I. Since the dimensions of $\mathcal{R}_t$ and $\mathcal{Y}_t$ are identical, high-dimensional tensor-based monitoring methods are proposed. For example, the multilinear principal component analysis (MPCA) method can be employed for feature extraction (Yan et al., 2014), we have the following optimization problems in phase I.

$$\min_{\mathcal{C}_\mathcal{R}, U_1, U_2, U_3} \|\mathcal{R} - \mathcal{C}_\mathcal{R} \times_2 \boldsymbol{U}_1 \times_3 \boldsymbol{U}_2 \times_4 \boldsymbol{U}_3\|$$

$$\text{s.t. } \boldsymbol{U}_i^T \boldsymbol{U}_i = \boldsymbol{I} \text{ for } i \in \{1,2,3\}, \quad (12)$$

where $\mathcal{R} \in \mathbb{R}^{n \times Q_1 \times Q_2 \times Q_3}$ is the set of $\mathcal{R}_t$, and $\mathcal{C}_\mathcal{R}$ is its core tensor.

Based on the extracted features, different types of control charts can be applied for monitoring. For example, we can build $T^2$ and $Q$ charts to monitor $\mathcal{C}_\mathcal{R}$ and projection residuals $e = \mathcal{R} - \mathcal{C}_\mathcal{R} \times_1 \boldsymbol{U}_1 \times_2 \boldsymbol{U}_2 \times_3 \boldsymbol{U}_3$, respectively. In phase II, the online monitoring statistics are calculated as $\mathcal{C}_{\mathcal{R}_{new}} = \mathcal{R}_{new} \times_1 \boldsymbol{U}_1^T \times_2 \boldsymbol{U}_2^T \times_3 \boldsymbol{U}_3^T$ and $e_{new} = \mathcal{R}_{new} - \mathcal{C}_{\mathcal{R}_{new}} \times_1 \boldsymbol{U}_1 \times_2 \boldsymbol{U}_2 \times_3 \boldsymbol{U}_3$. Similarly, statistics in EWMA charts can also be extended in tensor space as $Z_t = \omega Z_{t-1} + (1-\omega)\text{vec}(\mathcal{C}_\mathcal{R})$, to monitor subtle variations and drifts in extracted features of $\mathcal{R}_t$. Appendix D presents more details, and numerical studies in Section 5 validate the performances of the proposed monitoring methods.

## 4. Simulation for performance evaluation

In this section, we evaluate two parameter estimation algorithms proposed in Section 3.2 across different scenarios and confirm the stability region of the EWMA controller in Section 3.3. Section 4.1 introduces the details of data generation process. In Section 4.2, the accuracy of two algorithms for tensor parameter estimation is analyzed. Finally, Section 4.3 is dedicated to verify the performance of the EWMA controller.

### *4.1 Data generation*

In simulation cases, we consider 6 root causes, which correspond to 6 dimensions of control recipes at each run ($\boldsymbol{u}_t \in \mathbb{R}^6$). The dimension of the output image at each run is $100 \times 200 \times 2$, i.e., $\mathcal{Y}_t \in \mathbb{R}^{100\times200\times2}$, where 2 denotes the two components of overlay errors in *x*-axis and *y*-axis, respectively. This means that each wafer contains 20,000 sensor points. The corresponding tensor parameter is $\mathcal{B} \in \mathbb{R}^{6\times100\times200\times2}$. Suppose that there are 10 offline manufacturing cycles in offline dataset, with $T = 30$ runs in each cycle (i.e., $n = 300$). The desired target of the overlay error is set as $\mathcal{Y}^* = \mathbf{0} \in \mathbb{R}^{100\times200\times2}$.

We first generate the tensor parameter and input control recipes. For input recipes, we randomly simulate their elements from a multinormal distribution $N(0,1)$. Suppose that $P_1 = 2$, $P_2 = 3$, and $P_3 = 2$, which means that the core tensor of $\mathcal{Y}_t$ is $\mathcal{C}_{\mathcal{Y}_t} \in \mathbb{R}^{2\times3\times2}$. Therefore, the corresponding parameters can be generated by $\mathcal{B} = \mathcal{C}_{\mathcal{B}} \times_2 \boldsymbol{V}_1 \times_3 \boldsymbol{V}_2 \times_4 \boldsymbol{V}_3$, where $\mathcal{C}_{\mathcal{B}} \in \mathbb{R}^{6\times2\times3\times2}$. To verify Algorithms 1 and 2 proposed in Section 3.2, we generate a core tensor based on the condition that mode-1 unfolding of $\mathcal{C}_{\mathcal{B}}$ is row-sparsity.

Furthermore, according to Zhong et al. (2023), we set $\boldsymbol{V}_i$ $(i = 1,2)$ as

$$\boldsymbol{V}_i = \left[\sin(\pi i x_{i1}), \ldots, \sin(\pi i x_{iQ_i})\right]^T \tag{13}$$

where $x_{ij} = \frac{j}{Q_i}$, and $j \in \{1,2, \ldots Q_i\}$. Considering the significant impact of the correlation between disturbances and control recipes on Algorithm 1, we discuss different correlations, which are characterized by correlation matrix $\boldsymbol{A}$. $\boldsymbol{A} = a * \boldsymbol{I}_{6\times6}$, and $a$ equals -50, -20, 0, 20, and 50, respectively. Following Equation (6), we have $\boldsymbol{d}_t = \boldsymbol{A}^T\boldsymbol{u}_t + \boldsymbol{W}_t$, where $\boldsymbol{W}_t$ is generated from the standard normal distribution. Finally, based on $\boldsymbol{u}_t$, $\boldsymbol{d}_t$, and $\mathcal{B}$, we simulate the response tensor $\mathcal{Y}_t$ based on the process model in Equation (1). Furthermore, $\mathcal{E}_t$ represents the second type of disturbance defined as random noises in offline data, whose elements are sampled from a standard normal distribution. Figure 3 illustrates a simulation sample based on different correlations between $\boldsymbol{u}_t$ and $\boldsymbol{d}_t$.

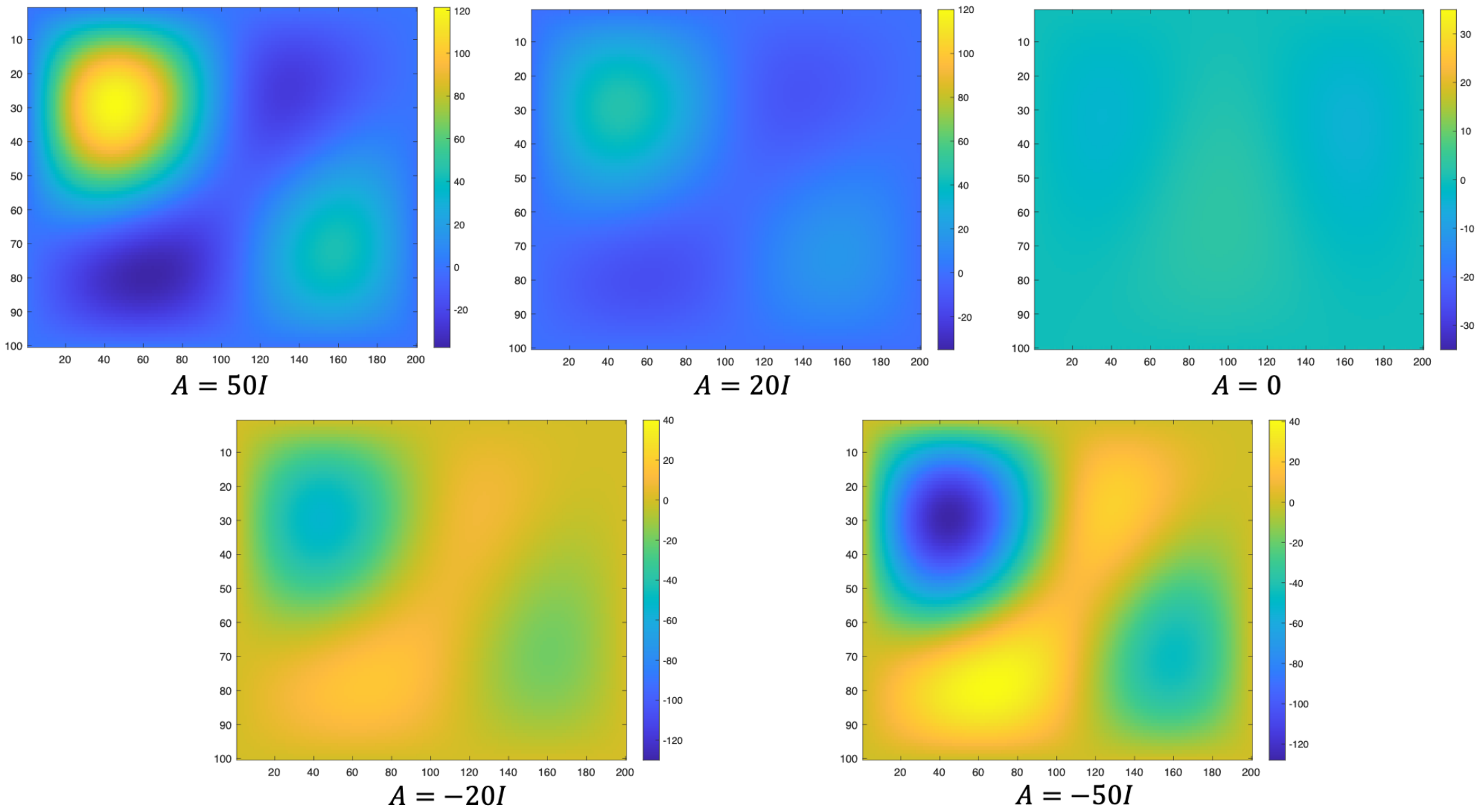

Figure 3. Simulation results based on different correlations between $\boldsymbol{u}_t$ and $\boldsymbol{d}_t$.

### *4.2. Performance of parameter estimation algorithms*

Before control optimization, it is crucial to accurately estimate the high-dimensional parameters in the process model. Based on the practical applications in lithography processes, we can simplify the tensor parameter according to its fourth order, and divide it into $\mathcal{B}_1$ and $\mathcal{B}_2$ ($\mathcal{B}_1, \mathcal{B}_2 \in \mathbb{R}^{m \times Q_1 \times Q_2}$), which represent the coefficients corresponding to the component of overlay errors in the horizontal and vertical directions, respectively. Suppose that the errors in these two directions are independent, we take the estimation method of parameter $\mathcal{B}_1$ as an example to analyze the performance of different parameter estimation algorithms. Estimation for $\mathcal{B}_2$ follows the same logic.

To evaluate the accuracy of parameter estimation in the process model, we use the parameter estimation errors defined as $PEE = \left\|\text{vec}(\mathcal{B} - \widehat{\mathcal{B}})\right\|_F$ to measure the performance of two algorithms in different correlations (i.e., $\boldsymbol{A}$) between $\boldsymbol{u}_t$ and $\boldsymbol{d}_t$. We use the same offline dataset with data size $n = 300$ to compare the performance of Algorithms 1 and 2, and the results of PEE with 50 replications are presented in Figure 4. More details about the parameter adjustment are presented in Appendix E.

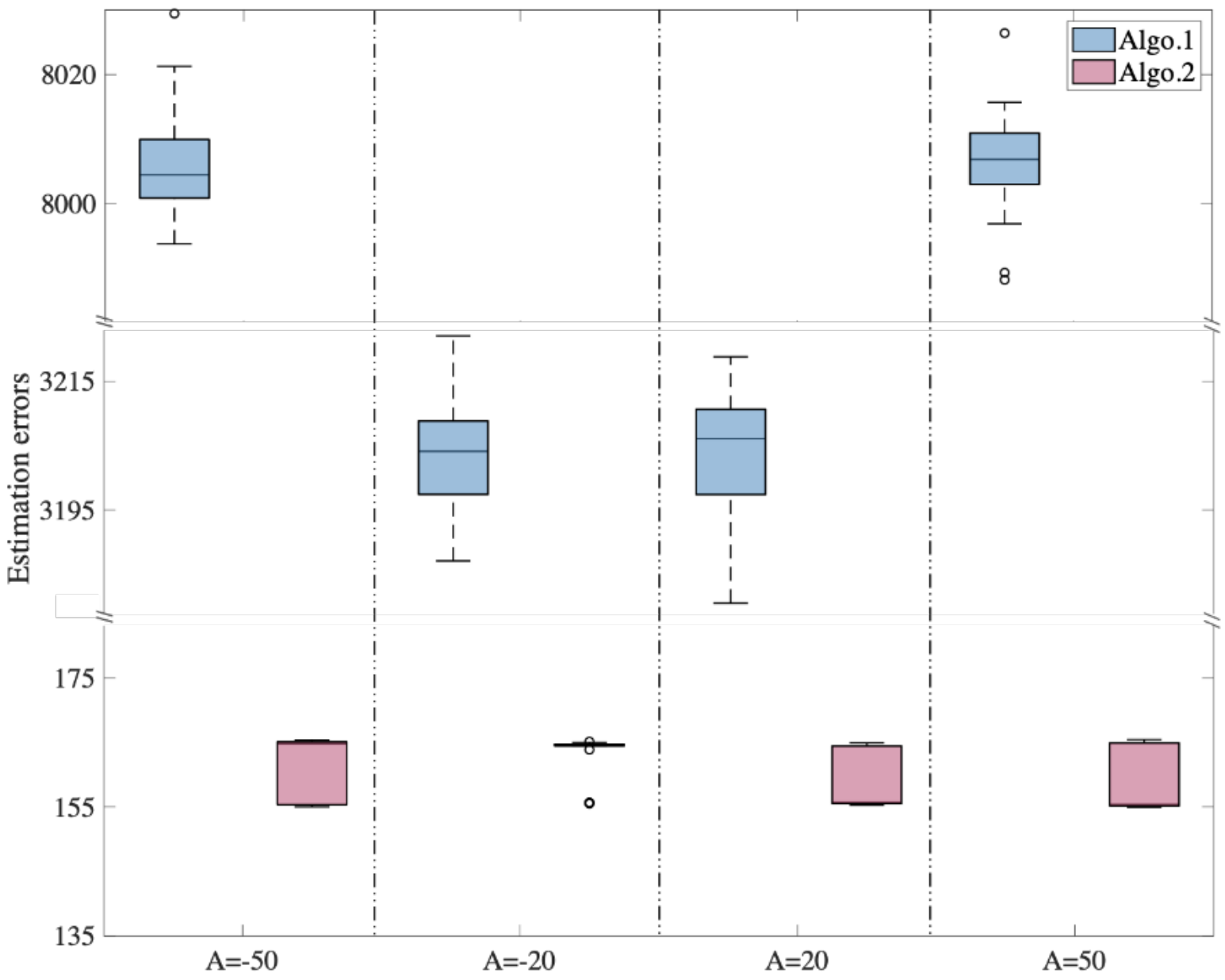

Figure 4. PEE of tensor parameter estimators of different algorithms

As shown in Figure 4, it is observed that the performance of Algorithm 1 is sensitive to the correlations between disturbances ($\boldsymbol{d}_t$) and control recipes ($\boldsymbol{u}_t$), and the estimation errors are increased with correlations between $\boldsymbol{d}_t$ and $\boldsymbol{u}_t$. In contrast, Algorithm 2 exhibits more stable performances when $\mathcal{B}$ is row-sparse, regardless of how the correlation between $\boldsymbol{d}_t$ and $\boldsymbol{u}_t$ changes.

### *4.3. Stability verification of EWMA controller in tensor space*

To verify the stability of the EWMA controller based on biased parameter estimators, we use Algorithm 1 to estimate the tensor parameter especially when disturbances are correlated with control recipes. Then, the performance of the EWMA controller with different tuning parameters is discussed based on tensor parameter estimators with different errors. We use the mean absolute error (MAE) for each batch (i.e., $\frac{1}{T}\sum_{t=1}^{T}\|\mathcal{Y}_t\|_F$) to evaluate the control performance, which is presented in Table 1.

Table 1: Summary of MAE with different parameter estimators

| MAE | **$\lambda$=0.1** | **$\lambda$=0.3** | **$\lambda$=0.5** | **$\lambda$=0.7** | **$\lambda$=0.9** |
|---|---|---|---|---|---|
| $\frac{1}{T}\sum_{t=1}^{T}\|\mathcal{Y}_t\|_F$ | **Δ=-0.95** | **Δ=-0.85** | **Δ=-0.75** | **Δ=-0.65** | **Δ=-0.55** |
| $A=-0.9\cdot\boldsymbol{I}$ | 0.2320 | $\mathbf{6.2789\times 10^{10}}$ | $\mathbf{8.9286\times 10^{18}}$ | $\mathbf{7.2738\times 10^{23}}$ | $\mathbf{2.4359\times 10^{27}}$ |
| $A=-0.6\cdot\boldsymbol{I}$ | 0.1749 | 0.1510 | 0.1820 | **0.3602** | **311.6865** |
| $A=-0.3\cdot\boldsymbol{I}$ | 0.1921 | 0.1510 | 0.1439 | 0.1513 | 0.1734 |
| $A=0$ | 0.2057 | 0.1600 | 0.1452 | 0.1415 | 0.1446 |
| $A=10\cdot\boldsymbol{I}$ | 0.2843 | 0.2568 | 0.2373 | 0.2227 | 0.2113 |
| Benchmark: error in the case that without control: 0.3025 | | | | | |

In Table 1, we provide a summary of the MAE values, which are based on varying correlation values ($\boldsymbol{A}$) between disturbance and control recipes, as well as tuning parameters ($\lambda$) for EWMA controllers. We also present the values of $\Delta := (\lambda-2)/2$ and the eigenvalues of $A$ in Table 1 to verify Corollary 2. The case without control is defined as a benchmark, and the control costs outside the stability region are emphasized by bolding. Theoretically, the stable condition is $\text{eig}(\boldsymbol{A}) > (\lambda-2)/2$. However, the error terms $\boldsymbol{W}_t$ in data generation potentially induce bias in $\boldsymbol{A}$. In summary, the stability condition of the EWMA controller in tensor space has been confirmed for control image-based quality variables.

## 5. Case study

In this Section, we employ the proposed framework for parameter estimation, control optimization, and monitoring methods in a lithography process, which is a crucial stage during the semiconductor manufacturing process. The objective of control within the lithography process is to minimize the overlay errors among various layers of photoresist materials. The data utilized in this section has been authenticated by Armitage and Kirk (1988) and Brunner et al. (2013), and extensively applied in studies by Gahrooei et al. (2021) and Zhong et al. (2021). To validate our method, we employ the same simulator as Zhong et al. (2023), with the details presented comprehensively in their appendix.

### *5.1 Comparison of different process controllers*

In literature, a recent feedback image-based controller is proposed by Zhong et al. (2023), which is defined as the basic image-based controller and considered as a benchmark in this case study. For fair comparisons, we generate the same offline data and use Algorithm 1 for parameter estimation in both two controllers.

In the setting of Zhong et al. (2023), disturbances are perceived as random errors and their autocorrelations are not taken into account. The control scheme is designed as $u_t = \boldsymbol{C}_{\mathcal{B}(1)}^{-1}\left(\widehat{\boldsymbol{V}}_{\mathbf{3}} \otimes \widehat{\boldsymbol{V}}_{\mathbf{2}} \otimes \widehat{\boldsymbol{V}}_{\mathbf{1}}\right)^T vec(R_{t-1})$, where $R_{t-1} = \mathcal{Y}_{t-1} - \mathcal{Y}^*$. To comprehend the necessity of integrating autocorrelations in disturbances, we use integrated moving average (IMA) and autoregressive integrated moving average (ARIMA) processes for disturbance description, which is widely accepted in semiconductor manufacturing (He et al., 2009; Liu et al., 2018).

Tables 2 and 3 illustrate performance comparisons between the basic image-based controller, the proposed EWMA controller with a tuning parameter $\lambda = 0.5$, and the case without control based on the IMA and ARIMA disturbances, respectively. The results reveal that the proposed EWMA controller outperforms the other two control schemes in compensating for the IMA and ARIMA disturbance processes. In comparison with Zhong et al.'s controller, the proposed EWMA control scheme offers superior accuracy in predicting unstable disturbances with complex autocorrelations.

Table 2. Performances based on IMA disturbances

| MAE $(\frac{1}{T}\sum_{t=1}^{T}\|\mathcal{Y}_t\|_F)$ | $\theta = 0.3$ | $\theta = 0.5$ | $\theta = 0.7$ |
|---|---|---|---|
| Our controller (EWMA) | 0.1424 | 0.1379 | 0.1407 |
| Zhong et al.'s controller | 0.3265 | 0.2614 | 0.1886 |
| Without control | 0.3443 | 0.3418 | 0.3670 |

Table 3. Performances based on ARIMA disturbance

| MAE $(\frac{1}{T}\sum_{t=1}^{T}\|\mathcal{Y}_t\|_F)$ | $\varphi = 0.25$ | | | $\varphi = 0.75$ | | |
|---|---|---|---|---|---|---|
| | $\theta = 0.3$ | $\theta = 0.5$ | $\theta = 0.7$ | $\theta = 0.3$ | $\theta = 0.5$ | $\theta = 0.7$ |
| Our controller (EWMA) | 0.1567 | 0.1430 | 0.1385 | 0.2593 | 0.2075 | 0.1681 |
| Zhong et al.'s controller | 0.3198 | 0.3076 | 0.3109 | 0.6189 | 0.4751 | 0.3506 |
| Without control | 0.4469 | 0.3335 | 0.2292 | 1.1747 | 0.8582 | 0.5499 |

### *5.2 Performance analysis of monitoring methods*

In the aforementioned analysis presented in Section 5.1, the second type of disturbance $\mathcal{E}_t$ is assumed as random error from manufacturing environments, and cannot be compensated by control recipes. However, if there are random shocks or irreversible drifts in the environments, $\mathcal{E}_t$ may no longer be stable and independent. We propose the monitoring methods for these cases. When the monitoring statistics exceed their control limit, a control chart warning ensues, necessitating the suspension of the production line for further inspection.

To analyze different types of drifts in disturbance $\mathcal{E}_t$, we first define the in-control (IC) state and four types of out-of-control (OC) patterns. Suppose that in an IC manufacturing process, the elements in $\mathcal{E}_t$ follow an independent and identical normal distribution $\mathcal{N}(0, \sigma_0^2)$, while different OC cases are presented as follows:

- Case A: Mean shift, i.e., $\mathcal{E}_t \sim \mathcal{N}(\mu, \sigma_0^2)$, where $\mu \neq 0$.

- Case B: Variation shift, i.e., $\mathcal{E}_t \sim \mathcal{N}(0, \sigma_1^2)$, where $\sigma_1 = 2 \cdot \sigma_0$.
- Case C: Gradual drift with IMA process, i.e., $\mathcal{E}_t \sim \mathrm{IMA}(1,1)$.
- Case D: Gradual drift with ARIMA process, i.e., $\mathcal{E}_t \sim \mathrm{ARIMA}(1,1,1)$.

OC patterns in Cases A and B usually exhibit large shifts, which are easy to monitor. We use $T^2$ and $Q$ charts proposed in Section 3.4 to monitor the OC patterns in Cases A and B. Due to the significant shifts, even based on different feedback control schemes, $T^2$ and $Q$ charts can efficiently identify the OC patterns on $\mathcal{E}_t$. Suppose that a shift or drift occurred at run $t = 21$, the performance of control charts in Cases A and B is illustrated in Figure 5(a) and (b), respectively. As shown, it is obvious that the $T^2$ chart demonstrates significant sensitivity in mean and variance shifts, while $Q$ chart can efficiently detect changes in variance. Consequently, if the distribution of $\mathcal{E}_t$ changes significantly, combining $T^2$ and $Q$ charts is a feasible monitoring scheme.

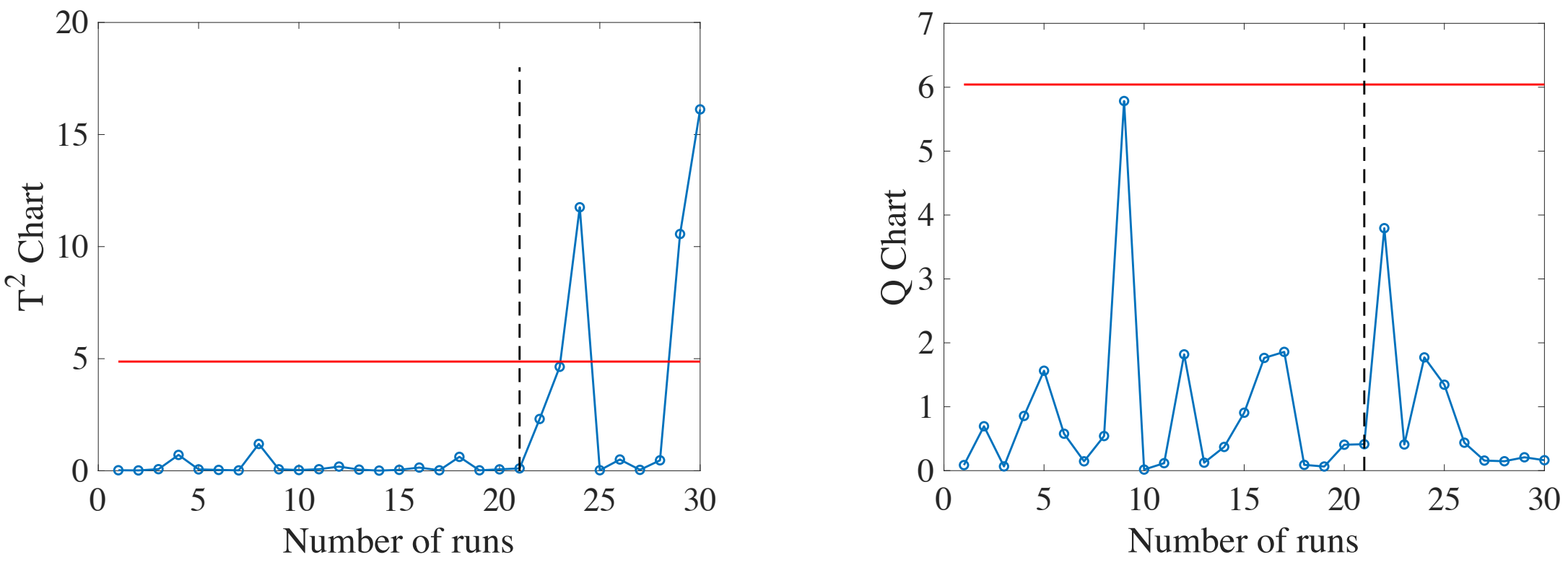

Figure 5 (a). $T^2$ and $Q$ charts in Case A

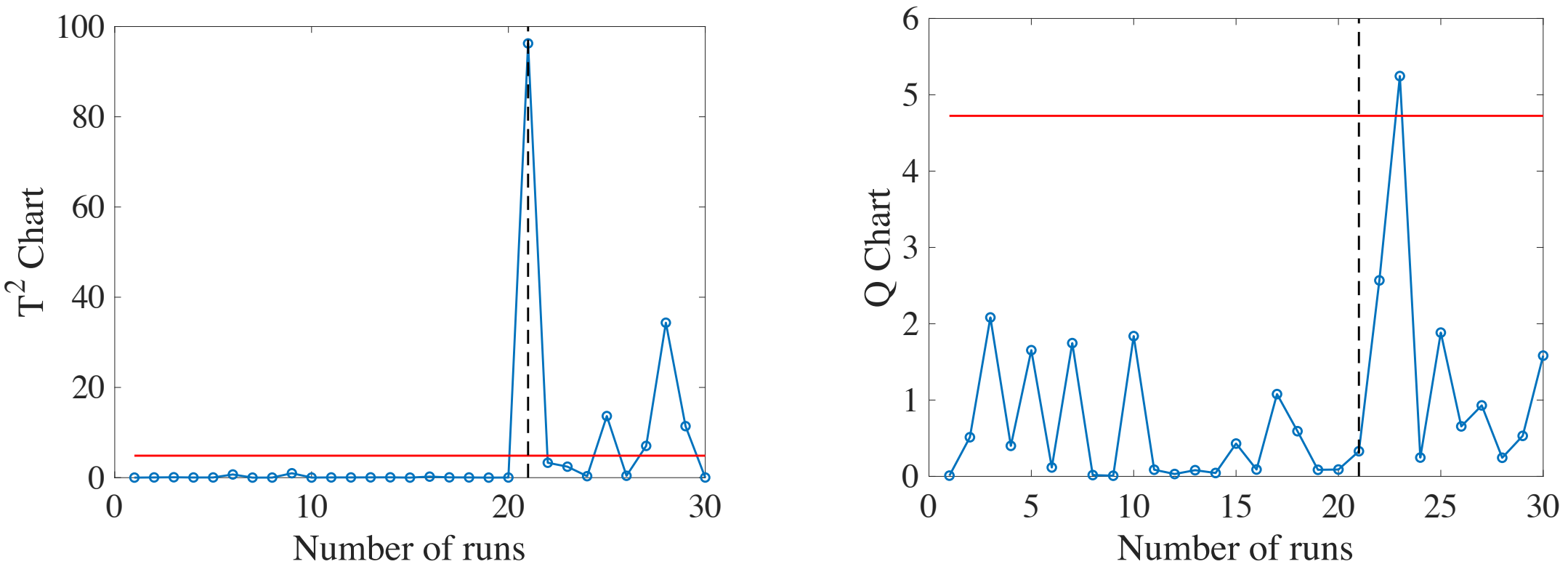

Figure 5 (b). $T^2$ and $Q$ charts in Case B

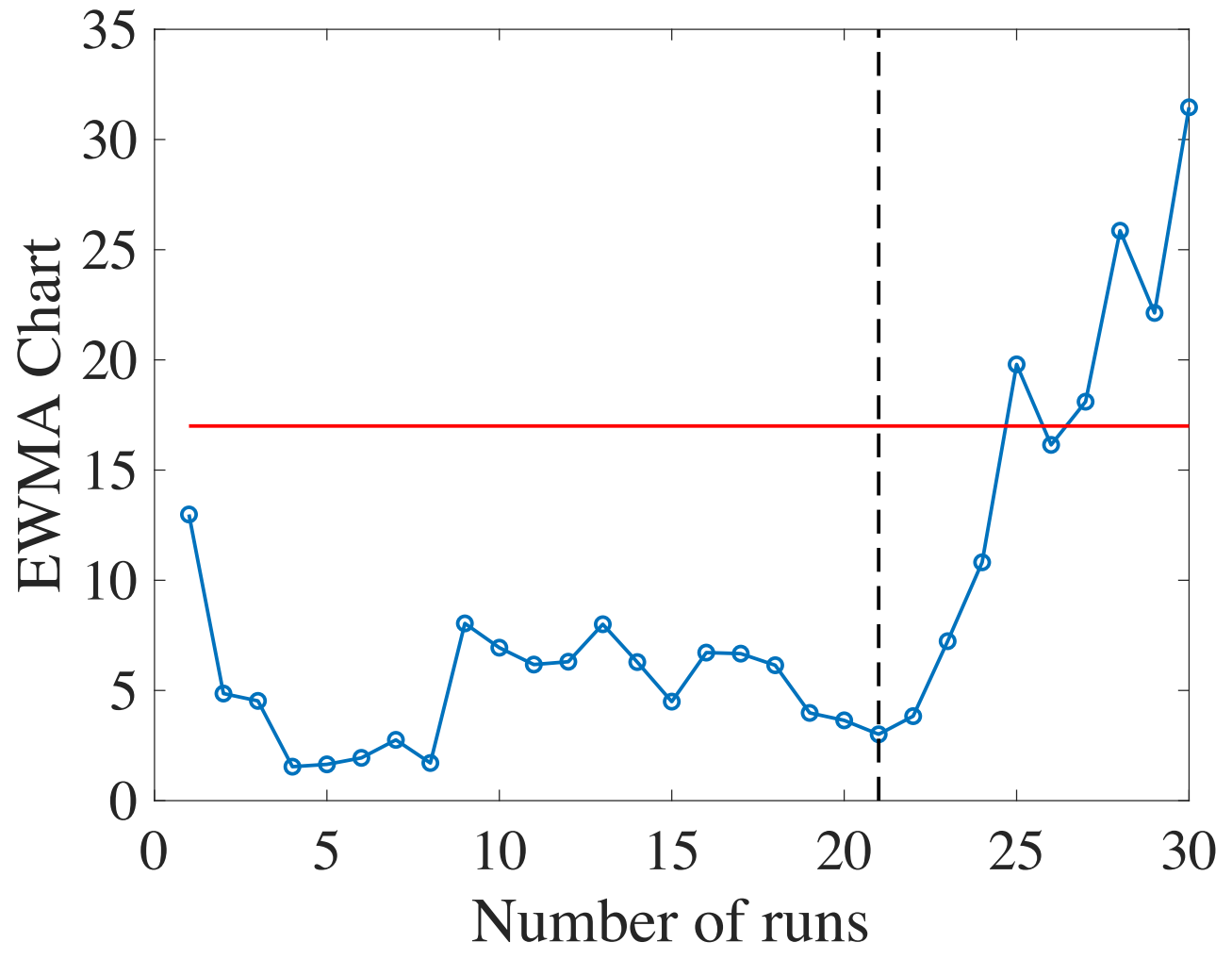

Figure 5 (c). EWMA charts in Case C

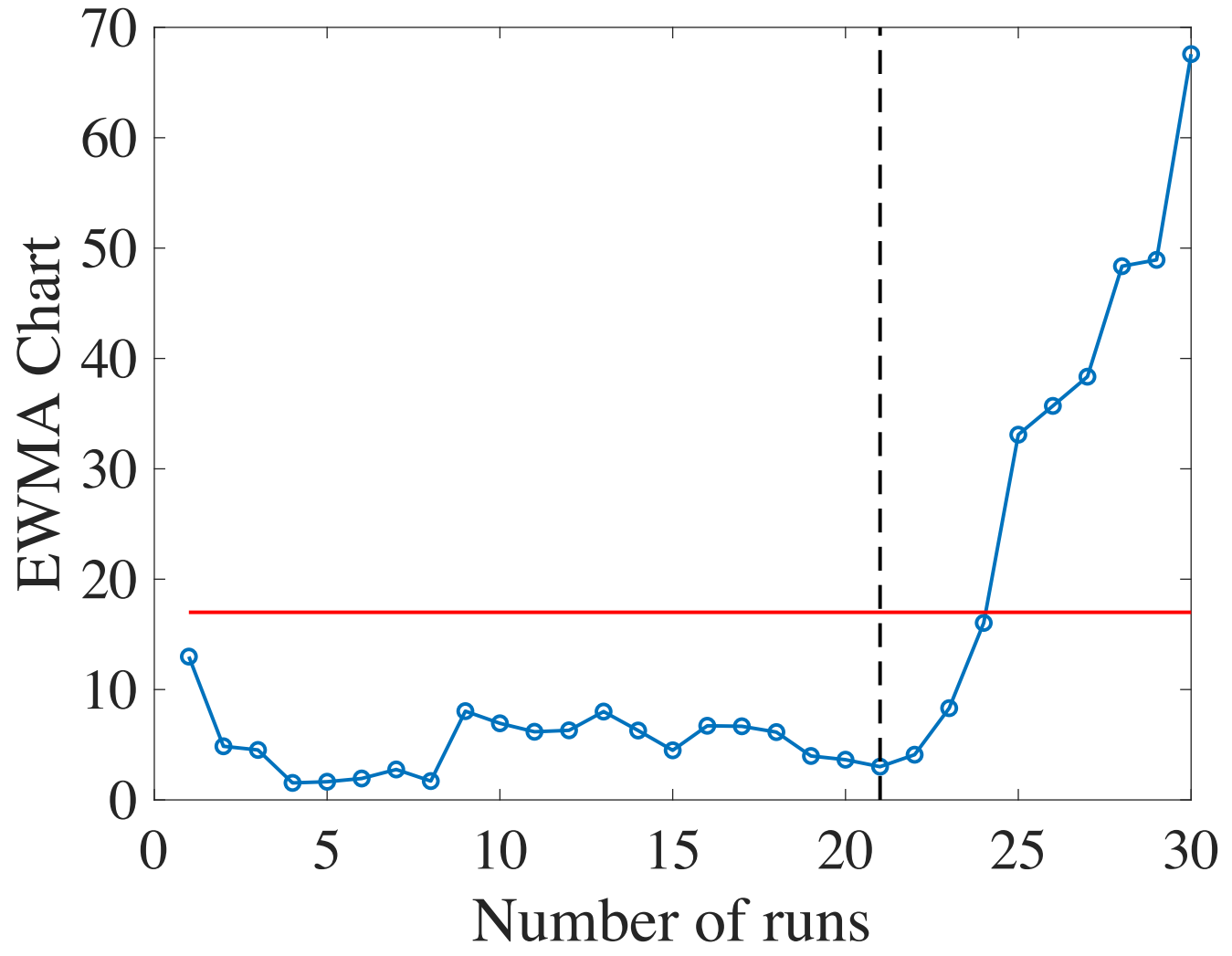

Figure 5 (d). EWMA charts in Case D

As $T^2$ charts lack sensitivity when detecting minor shifts, we extend the EWMA control chart into tensor space to detect the gradual drifts in Cases C and D. As shown in Figures 5 (c) and (d), if $\mathcal{E}_t$ gradually drifts according to a specific time series such as the IMA or ARIMA process, the EWMA chart can be extended into tensor space to detect this drift. However, EWMA charts are more sensitive to process changes, control recipes can also affect the monitoring results. Interestingly, we observed that ARIMA processes have larger drift magnitudes than IMA processes, thereby facilitating the change detections by the EWMA chart.

## 6. Conclusion

With the popularity of sensor technology, quality measurement is not restricted to univariate or multivariate variables but involves image-based quality variables. In this work, we focus on process control and monitoring scheme design using tensors for lithography processes with image-based quality variables. Given the limited number of input control variables, we categorize disturbances into two types based on whether they can be compensated by control recipes. These two types of disturbances increase the complexity of high-dimensional tensor parameter estimation, making identifiable parameter estimation and process controller design quite challenging.

To enhance the precision of parameter estimators for process control, we improve existing least square estimation algorithms in tensor space by sparse learning to improve its limitations, especially when control recipes and the first type of disturbance in the offline dataset are highly correlated. Subsequently, the EWMA control scheme is extended into tensor space to compensate for the first type of disturbance, and different monitoring methodologies are introduced for the second type of disturbance. Importantly, the properties of parameter estimation algorithms and the stability of the EWMA controller are guaranteed. To illustrate the efficiency of the proposed parameter estimation, process control, and monitoring methods, we propose extensive simulations and case studies for verification. By comparing with the existing image-based process controller, we discovered that the EWMA controller outperforms in predicting and compensating for unstable disturbances with autocorrelations. Finally, $T^2$, $Q$, and EWMA charts are also employed for efficient monitoring of different types of variations in the second type of disturbance arising from manufacturing environments, establishing a comprehensive process control and monitoring methodology.

For future works, more control schemes such as double-EWMA and variable-EWMA controllers can also be extended in tensor spaces for more accurate control. Besides, spatial correlations on wafers could be incorporated into a process model.

# Appendix

## *Appendix A. Proof of Proposition 1*

According to the process model in Equation (1): $\mathcal{Y}_t = (\boldsymbol{u}_t + \boldsymbol{d}_t) * \mathcal{B} + \mathcal{E}_t$ and the regression formulation in Equation (6): $\boldsymbol{d}_t = \boldsymbol{A}^T \boldsymbol{u}_t + \boldsymbol{W}_t$, we can reformulate the process model as

$$\begin{aligned} \mathcal{Y}_t &= (\boldsymbol{u}_t + \boldsymbol{A}^T \boldsymbol{u}_t + \boldsymbol{W}_t) * \mathcal{B} + \mathcal{E}_t \\ &= \boldsymbol{u}_t(\boldsymbol{I}_m + \boldsymbol{A}) * \mathcal{B} + \boldsymbol{W}_t * \mathcal{B} + \mathcal{E}_t. \end{aligned}$$

As $\boldsymbol{W}_t$ and $\mathcal{E}_t$ are independent with $\boldsymbol{u}_t$, we define $\epsilon_t \coloneqq \boldsymbol{W}_t * \mathcal{B} + \mathcal{E}_t$ as random error that is independent of $\boldsymbol{u}_t$.

In Algorithm 1, we estimate the tensor parameter $\mathcal{B}$ based on $\mathcal{Y}_t$ and $\boldsymbol{u}_t$ by least square, when sample size reaches infinite, the estimator is unbiased for the parameter $(\boldsymbol{I}_m + \boldsymbol{A}) * \mathcal{B}$. Specifically, only when $\boldsymbol{A} = \boldsymbol{0}$, the LS estimation algorithm in tensor space is unbiased.

## *Appendix B. Proof of Theorem 1*

As analyzed in the fourth and fifth constraints of Equation (10), we have:

$$\hat{d}_{t+1} = \lambda\left(\mathcal{C}_{\mathcal{Y}_t} - \hat{\mathcal{C}}_{\mathcal{B}} \times_1 \boldsymbol{u}_t\right) + (1-\lambda)\hat{d}_t, \tag{B.1}$$

$$\boldsymbol{u}_{t+1} = -\left(\widehat{\boldsymbol{C}}_{\mathcal{B}(1)}^T \widehat{\boldsymbol{C}}_{\mathcal{B}(1)}\right)^{-1} \widehat{\boldsymbol{C}}_{\mathcal{B}(1)}^T * \mathrm{vec}(\hat{d}_{t+1}). \tag{B.2}$$

Based on the EWMA disturbance estimate in Equation (B.1), by substituting Equation (B.2) and the true process model in Equation (3.1), we have

$$\hat{d}_{t+1} = (I - \Lambda\boldsymbol{\xi})\hat{d}_t + \Lambda\left((\mathcal{Y}_t - \boldsymbol{u}_t \times_1 \mathcal{B}) \times_1 \hat{V}_1^T \times_2 \hat{V}_2^T \times_3 \dots \times_d \hat{V}_d^T\right), \tag{B.3}$$

where $\boldsymbol{\xi} = \left(\mathcal{B} \times_2 \hat{V}_1^T \times_3 \hat{V}_2^T \times_3 \dots \times_{d+1} \hat{V}_d^T\right)_{(1)} \left(\widehat{\boldsymbol{C}}_{\mathcal{B}(1)}^T \widehat{\boldsymbol{C}}_{\mathcal{B}(1)}\right)^{-1} \widehat{\boldsymbol{C}}_{\mathcal{B}(1)}^T$ and $\Lambda = \lambda I$ . To guarantee the stability of the control scheme, it is necessary to ensure that the sequence $\{\hat{d}_{t+1}\}$ does not diverge. Therefore, if we let $M = I - \Lambda\boldsymbol{\xi}$, and all eigenvalues of $M$ are less than 1, the EWMA controller in tensor space is stable. Specifically, if eigenvalues of the matrix $\boldsymbol{\xi}$ can be represented as $a_j + b_j i$, the stability condition can be expressed as: $\frac{2a_j}{a_j^2 + b_j^2} > \lambda$.

## *Appendix C. Proof of Corollary 2*

Specifically, if Algorithm 1 is used to estimate the tensor parameter in the process model, we have the condition $E(\widehat{\mathcal{B}}) = (I + A^*) \times_1 \mathcal{B}$ according to Theorem 1. Then, we can calculate the matrix $\boldsymbol{M} = I - \boldsymbol{\Lambda\xi} = I - \boldsymbol{\Lambda}(I + \boldsymbol{A})^{-1} \boldsymbol{C}_{\mathcal{B}(1)} \left(\widehat{\boldsymbol{C}}_{\mathcal{B}(1)}^T \widehat{\boldsymbol{C}}_{\mathcal{B}(1)}\right)^{-1} \widehat{\boldsymbol{C}}_{\mathcal{B}(1)}^T$. Therefore, the stability region $|\text{eig}(\boldsymbol{M})| < 1$ is equivalent to

$$0 < \text{eig}(I + \boldsymbol{A})^{-1} < \frac{2}{\lambda}. \tag{C.1}$$

Equation (C.1) can be further simplified as $\text{eig}(\boldsymbol{A}) > \frac{\lambda - 2}{2}$, which is presented in Corollary 2.

## *Appendix D. More details about process monitoring*

In Section 3.4, we introduce three kinds of control charts in tensor space for Type II disturbance monitoring, and we present these three charts respectively.

We use $T^2$ chart to monitor the elements in core tensor of the control residuals (i.e., $\mathcal{R}$). First, the features in core tensor are extracted by Tucker decomposition. Then, by transforming the core tensor into a vector with dimension $W \times 1$ ($W = P_1 \times P_2 = 6$ in our case study), we have the monitoring statistics for online monitoring:

$$T_{new}^2 = (C_{\mathcal{R}new} - \bar{C}_{\mathcal{R}}) S_R^{-1} (C_{\mathcal{R}new} - \bar{C}_{\mathcal{R}}) * \frac{n(n-W)}{W(n^2-1)}, \tag{D.1}$$

where $S_{\mathcal{R}}$ and $\bar{C}_{\mathcal{R}}$ are the covariance matrix and mean vector of $\text{vec}(\mathcal{C}_{\mathcal{R}})$ estimated by offline data, respectively, $C_{\mathcal{R}new}$ is the online observation, and $n$ is the total number of offline data.

$Q$ chart is used to monitor the errors after the Tucker decomposition, i.e., $e = \mathcal{R} - \mathcal{C}_{\mathcal{R}} \times_1 \boldsymbol{U}_1 \times_2 \boldsymbol{U}_2 \times_3 \boldsymbol{U}_3$, and we have the online $Q$ statistics as $Q_{new} = \|e_{new}\|^2$. According to Nomikos and MacGregor (1995) and Yan et al. (2014), if extracted features and errors are normally distributed, $T_{new}^2$ follows $F$ distribution with degrees of freedom $W$ and $n - W$. $Q_{new}/g$ follows $\chi^2$ distribution with degrees of freedom $h$, where $g$ and $h$ can be solved by equations as follows: $\text{E}(Q_{new}) = gh$, $\text{var}(Q_{new}) = 2g^2h$. Similar to the multivariate $T^2$ and $Q$ control charts, the control limits can be determined by the $(1 - \alpha)100^{th}$ percentile. In our case study, $\alpha$ is selected as 0.025.

EWMA chart is used to monitor subtle variations of elements in core tensor of $\mathcal{C}_{\mathcal{R}_t}$. We vectorize of $\mathcal{C}_{\mathcal{R}_t}$ and propose the monitoring statistics as: $Z_t = (1 - \omega) Z_{t-1} + \omega \text{vec}(\mathcal{C}_{\mathcal{R}_t})$, with mean and covariance matrix equal to $\bar{Z}$ and $\boldsymbol{\Sigma}_{Z_t} = \left\{\frac{\omega[1-(1-w)^{2t}]}{2-\omega}\right\} \cdot \boldsymbol{\Sigma}_{\mathcal{R}_t}$ respectively, where $\mathbf{\Sigma}_{\mathcal{R}_t}$ is the covariance matrix of $\mathrm{vec}(\mathcal{C}_{\mathcal{R}_t})$. The control statistic is calculated as $T_Z^2 = Z_t'\mathbf{\Sigma}_{Z_t}^{-1}Z_t$, when $T_Z^2 > L$, the process is out of control. According to Linderman and Love (2000), $L$ is chosen to achieve a specified in-control ARL (i.e., $ARL_0$). In our case study, $ARL_0$ equals to 200.

## *Appendix E. Details about estimation algorithms.*

In Section 3.2, we propose a row-sparse assumption for the tensor parameter and obtain an identifiable estimator. Specifically, the tensor parameters and correlations are solved by the GLRP optimization problem. According to Bing et al. (2022) and Bing et al. (2023), the tuning parameters can be selected as follows.

We first select the value of $\lambda_2$ according to Figure E1 to guarantee the sparsity of the correlation matrix $\boldsymbol{A}$. In this case, we let $\lambda_2 = 2$.

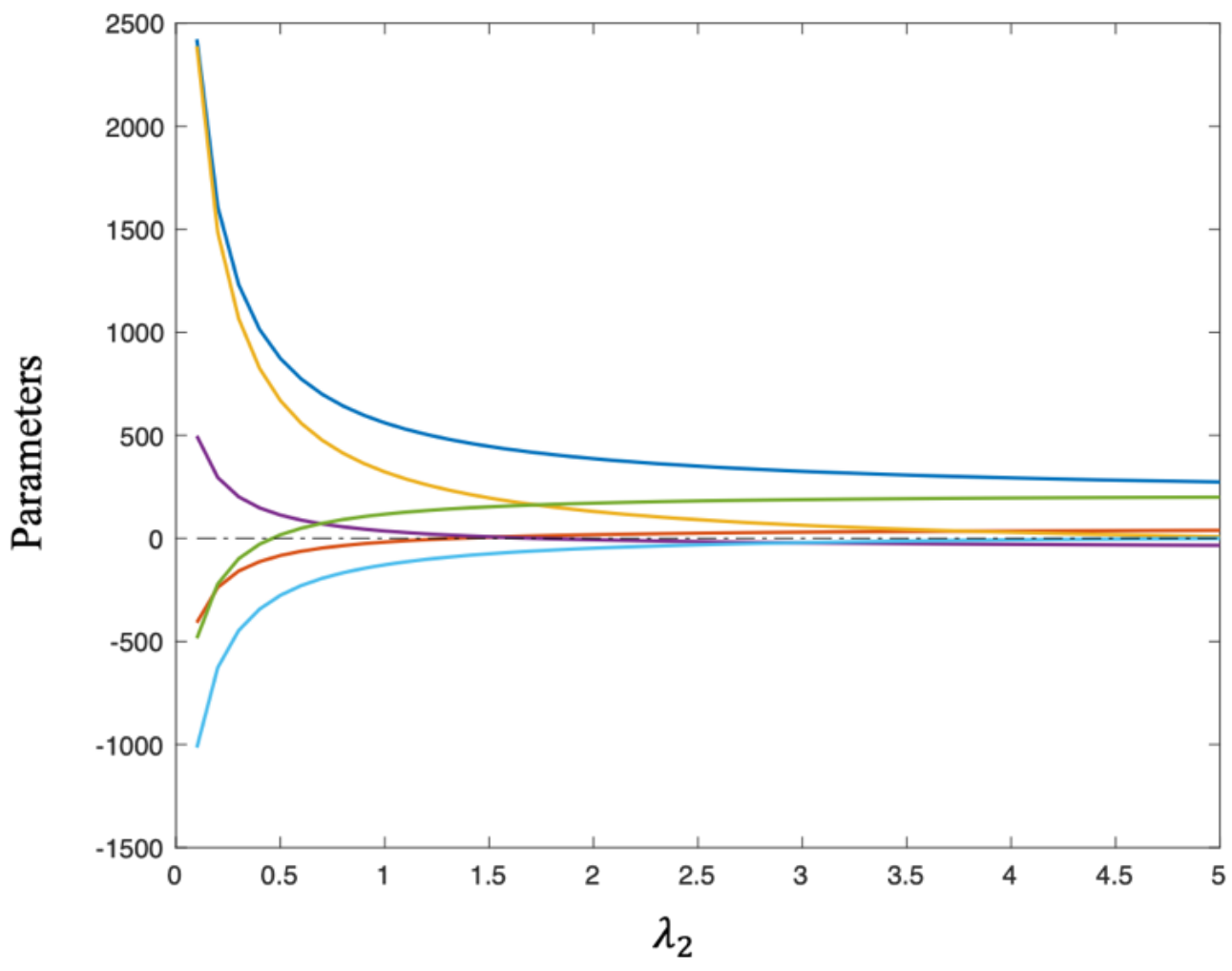

Figure E1. Illustration of parameters based on different $\lambda_2$

Then, we adopt the method in Bing et al. (2022) to select $\lambda_1$ given $\lambda_2$ in Equation (E.1).

$$\lambda_1(\lambda_2) = c_0\sqrt{\max_{1\le j\le p} M_{jj}(\lambda_2)}\left(\sqrt{\frac{m}{n}} + \sqrt{\frac{2\mathrm{log}p}{n}}\right), \qquad \text{(E.1)}$$

where $M(\lambda_2) = n^{-1}\boldsymbol{U}^T Q_{\lambda_2}^2 \boldsymbol{U}$, with $Q_{\lambda_2}^2 = I_n - \boldsymbol{U}\left(\boldsymbol{U}^T\boldsymbol{U} - n\lambda_2 I_p\right)^{-1}\boldsymbol{U}^T$. $\boldsymbol{U}$ is the set of control recipes in offline data, $p$ is the dimension of $\boldsymbol{U}$, $m$ is the dimension of core tensor for the output image, and $n$ is the total number of observations in the offline dataset. $c_0 = 1$ is a constant according to Bing et al. (2022).